\documentclass[pdflatex,sn-mathphys-num]{sn-jnl}
\usepackage{graphicx}%
\usepackage[table,xcdraw,usenames,dvipsnames]{xcolor}
\usepackage{multirow}%
\usepackage{amsmath,amssymb,amsfonts}%
\usepackage{amsthm}%
\usepackage{mathrsfs}%
\usepackage[title]{appendix}%
\usepackage{textcomp}%
\usepackage{manyfoot}%
\usepackage{booktabs}%
\usepackage{algorithm}%
\usepackage{algorithmicx}%
\usepackage{algpseudocode}%
\usepackage{listings}%
\usepackage{caption}
\usepackage{comment}
\usepackage{bm}
\usepackage{bbold}
\restylefloat{table}
\DeclareMathOperator*{\argmax}{arg\,max}

\begin{document}

\title[Ordinal Semantic Segmentation Applied to Medical and Odontological Images]{Ordinal Semantic Segmentation Applied to Medical and Odontological Images}

\author[1]{\fnm{Mariana Dória Prata} \sur{Lima}}\email{mdoria@posgrad.lncc.br}

\author[2]{\fnm{Gilson} \sur{Antonio Giraldi}}\email{gilson@lncc.br}                          
\author[3]{\fnm{Jaime} \sur{S. Cardoso}}\email{jaime.cardoso@inesctec.pt}

\affil[1, 2]{\orgname{National Laboratory for Scientific Computing}, \orgaddress{ \city{Petrópolis}, \state{Rio de Janeiro}, \country{Brazil}}}

\affil[3]{\orgname{INESC TEC, Faculty of Engineering, University of Porto}, \city{Porto}, \country{Portugal}}

\abstract{Semantic segmentation consists of assigning a semantic label to each pixel according to the corresponding classes. This process facilitates the understanding of object appearance and the spatial relationships among them, thus playing a fundamental role in the global interpretation of an image’s content and context. Although contemporary deep learning–based approaches have achieved high levels of accuracy, they often neglect ordinal relationships among classes, which may encode domain knowledge essential for correct scene interpretation. In this work, loss functions that incorporate such relationships into deep neural networks are investigated, promoting greater semantic consistency in semantic segmentation tasks. These loss functions are categorized as unimodal, quasi-unimodal, and spatial. Unimodal loss functions constrain the predicted probability distribution according to the class ordering, whereas quasi-unimodal loss functions relax this constraint by allowing small variations in the distribution shape while preserving coherence with the ordinal structure of the problem. In turn, spatial loss functions penalize semantic inconsistencies between neighboring pixels, favoring more coherent transitions in the image space. Specifically, this study adapts loss functions originally proposed for ordinal classification tasks to the ordinal semantic segmentation setting. Among them, the Expanded Mean Squared Error (EXP\_MSE), classified as unimodal, and the Quasi-Unimodal Loss (QUL), of quasi-unimodal nature, stand out, as well as the ordinal spatial loss function Contact Surface Loss using Signal Distance Function (CSSDF), which reinforces semantic consistency between adjacent pixels. These loss functions have been employed in medical imaging applications, where they have demonstrated promising performance, contributing to improved model robustness, enhanced generalization capability, and increased anatomical fidelity in the obtained results.}


\maketitle

\section{Introduction}

Deep learning based image segmentation has become one of the pillars of modern medical image analysis, enabling the automatic delineation of anatomical structures and pathological regions with unprecedented levels of accuracy~\cite{khan2021deep}. Despite these significant advances, deep learning–based approaches still face substantial challenges, such as a strong dependence on large annotated datasets and the high computational demands associated with the training process. Furthermore, a hypothesis widely discussed in the literature suggests that such networks lack intrinsic domain knowledge relevant to the task of semantic segmentation, which compromises their ability to correctly infer high-level semantic and structural relationships from the available training data.

For example, in several clinical scenarios, segmentation classes follow a natural order that reflects severity, depth, anatomical stratification, or disease progression. Examples include the grading of dermatological conditions such as eczema~\cite{sedai2017probabilistic}, the stratification of tumor infiltration levels, the segmentation of anatomical structures into layers~\cite{oktay2018anatomically}, and the organization of tissues along a depth axis. In such contexts, misclassifying a pixel (or voxel) into an adjacent ordinal class is clinically less critical than assigning it to a distant ordinal category. However, widely used loss functions and evaluation metrics, such as cross-entropy~\cite{shannon1948mathematical} and the Dice coefficient~\cite{taha2015metrics}, penalize classification errors symmetrically, explicitly disregarding this ordinal structure~\cite{sudre2017generalised,taha2015metrics}.

To address these limitations, a growing body of research has explored ordinal segmentation models that explicitly encode label ordering during both training and inference. These approaches are often inspired by ordinal regression~\cite{niu2016ordinal,Xintong2021,jenkinson2022universallyrankconsistentordinal}, ordinal depth estimation~\cite{zhao2017ordinal,chen2019ordinal}, hierarchical classification~\cite{ma2018hierarchical}, and probabilistic modeling~\cite{kendall2017uncertainties}. In the context of medical imaging, the incorporation of ordinal constraints has proven effective in enhancing model robustness and reducing anatomically implausible predictions, an aspect that is particularly relevant given the subjective nature and the recurring scarcity of medical annotations~\cite{jungo2018uncertainty,nair2020exploring}.

Several methods also exploit structural priors closely related to ordinality, including anatomical consistency, convexity, and depth awareness. Notable examples include neural networks with explicit anatomical constraints~\cite{oktay2018anatomically}, topology-preserving segmentation approaches~\cite{li2018topology}, and depth and convexity-aware optimization strategies developed to mitigate data scarcity, particularly in breast image segmentation tasks~\cite{zhao2022robust}.

Complementary research lines adopt probabilistic ordinal formulations to explicitly model the ambiguity associated with ordered labels and to quantify variability among human observers~\cite{sedai2017probabilistic,kendall2017uncertainties}.

Despite these advances, the literature on ordinal segmentation remains fragmented, with approaches spread across different research communities and characterized by heterogeneous terminology, modeling strategies, and evaluation protocols. Some proposals focus on redefining loss functions to penalize ordinal errors asymmetrically~\cite{chen2019ordinal}, while others rely on architectural constraints, label decomposition strategies, or iterative quality-aware regularization mechanisms~\cite{kervadec2019boundary,zhang2021quality}. Moreover, there is still no consensus on evaluation metrics capable of faithfully reflecting ordinal performance in medical image segmentation tasks, which hinders fair comparisons across different methods~\cite{taha2015metrics,cardoso2022metrics}.

Just focusing on probabilistic approaches, recently, several works have promoted or forced the output distribution to be
unimodal. This can be achieved either through the model's architecture or by using an appropriate loss during training~\cite{cardoso2022metrics}. While parametric models restrict their outputs to discrete unimodal probability distributions, parametric losses have also been used as an alternative to one-hot encoding for the target label distribution. Instead of assuming an a-priori probability distribution, non-parametric models have been proposed to ensure unimodality when converting the logits to probabilities. Non-parametric losses promote the neural network’s output probabilities to follow a unimodal distribution by imposing a set of constraints on all pairs of consecutive
labels, allowing for a more flexible decision boundary than parametric approaches. These approaches have been gener-
alized for quasi-unimodal distributions~\cite{qul22}.


In the context of ordinal semantic segmentation, the existing literature is much scarcer. 
Cruz et al.~\cite{Cruz_2025} formalize the task of ordinal segmentation in two consistency tasks: (i) representation consistency focuses on ordinal consistency within each individual pixel, and (ii) structural consistency focuses on ordinal consistency in the pixel grid, between neighbouring pixels. In particular, unimodal losses encourage the predicted probability distribution for each pixel to follow a unimodal shape centered around the ground-truth class, penalizing predictions that assign high probability to classes that are ordinally distant from the reference label; two new losses
are proposed for structural consistency.
On the other hand, quasi-unimodal losses were never considered before for ordinal semantic segmentation.


In this work, three ordinal loss functions are considered for the first time for ordinal segmentation, two of which were adapted in this study from loss functions originally proposed for ordinal classification. These formulations enable the explicit incorporation of ordinal knowledge into the training of deep neural networks without requiring significant architectural modifications, and are also naturally compatible with hybrid strategies that combine ordinal and categorical loss functions.

\begin{enumerate}

\item The \emph{Quasi-Unimodal Loss}, which imposes a quasi-unimodality constraint on the predicted probability distributions, encouraging the concentration of probability mass around the ground-truth label. This loss function was originally proposed in~\cite{qul22} for ordinal classification tasks and is adapted in this work to the context of ordinal semantic segmentation.

\item The \emph{EXP\_MSE Loss} was originally proposed in~\cite{expmse2026} for ordinal classification tasks and is adapted in this work to the ordinal segmentation setting. This loss function models the ordinal distance between classes and penalizes discrepancies between target and predicted distributions through the mean squared error.

\item The \emph{CSSDF Loss}, an adaptation of the spatial loss introduced in~\cite{Cruz_2025}, originally based on the Distance Transform, which is extended to employ a Signed Distance Function (\emph{SDF}). This modification aims to reinforce ordinal consistency between neighboring pixels and to preserve spatially coherent structural transitions in segmentation maps.

\end{enumerate}

The paper is organized as follows: Section~\ref{sec:mathematical-background} presents the main mathematical concepts, including a description of the loss functions used and the evaluation metrics; Section~\ref{sec:Experimental-evaluation} details the proposed method; Section~\ref{sec:results} presents and analyzes the experimental results; and Section~\ref{sec:conclusion} summarizes the main conclusions of the work.

\section{Mathematical Background}
\label{sec:mathematical-background}

In semantic segmentation, the learning problem is formulated as a dense labeling task over a spatial domain $\Omega \subset \mathbb{R}^d$, where each pixel (or voxel) $\mathbf{x} \in \Omega$ is assigned a class label. 
Given a labeled dataset $\mathcal{D} = \{(\mathbf{x}_i, y_i) ; i = 1, \dots, N\}$, each input image $\mathbf{x}_i \in \mathbb{R}^{H \times W \times B}$ is associated with a ground-truth label tensor $y_i \in \{1, \dots, K\}^{H \times W}$, where $K$ denotes the number of semantic classes.

From this dataset, a training subset $\mathcal{D}_{tr} \subset \mathcal{D}$, containing $N_{tr}$ image--mask pairs, is used to estimate the parameters of a neural network. The objective is to determine the optimal parameter vector $\mathbf{w}^* \in \mathbb{R}^{d}$, where $d$ is the number of trainable parameters, by minimizing a suitable loss function over $\mathcal{D}_{tr}$.

A deep neural network model is interpreted as a parameterized mapping
\[
f : \mathbb{R}^{H \times W \times B} \times \mathbb{R}^{d}
\longrightarrow
[0,1]^{H \times W \times K},
\]
Given an input image $\mathbf{x}_n$, the network produces
\[
f(\mathbf{x}_n; \mathbf{w}) = \hat{\mathbf{p}}_n \in [0,1]^{H \times W \times K},
\]
whose components $\hat{\mathbf{p}}_{n,i,j,k}$ represent the predicted probability that pixel $(i,j)$ in image $n$ belongs to class $k$ and $\mathbf{w}$ denotes the vector of trainable parameters.  The softmax activation ensures
\[
\sum_{k=1}^{K} \hat{\mathbf{p}}_{n,i,j,k} = 1
\quad \text{for all } n,i,j.
\]

In the classical formulation of semantic segmentation, the label set
\[
\mathcal{Y} = \{c_1, \ldots, c_K\}
\]
is treated as an unordered categorical set, meaning that all classes are considered mutually exclusive and independent, with no structural relationship among them.

Under this assumption, the most commonly used loss function is the categorical cross-entropy:
\begin{equation}
\mathcal{L}_{CE}(\mathcal{D}_{tr}; \mathbf{w}) 
=
- \frac{1}{N_{tr} \cdot H \cdot W} 
\sum_{n=1}^{N_{tr}} 
\sum_{i=1}^{H} 
\sum_{j=1}^{W} 
\sum_{k=1}^{K} 
\mathbb{1}(y_{n,i,j} = k) 
\log \hat{\mathbf{p}}_{n,i,j,k}.
\end{equation}

The training process consists of solving the optimization problem
\[
\mathbf{w}^* 
=
\arg\min_{\mathbf{w} \in \mathbb{R}^{d}}
\mathcal{L}_{CE}(\mathcal{D}_{tr}; \mathbf{w}),
\]
which corresponds to minimizing the empirical risk over the training dataset.

\vspace{0.5em}

While this formulation is appropriate for purely categorical labels, many segmentation problems involve ordered categories, where classes possess an inherent ranking. In such cases, ignoring the ordinal structure may lead to suboptimal learning, since misclassifying a pixel into a nearby class should be penalized less than assigning it to a distant one.

Ordinal segmentation can therefore be formally defined as a dense labeling problem in which each pixel (or voxel) $\mathbf{x} \in \Omega$ is associated with a label $y(\mathbf{x}) = c_k$ belonging to a totally ordered set
\[
\mathcal{Y} = \{c_1, \ldots, c_K\},
\quad \text{such that} \quad
c_1 \prec c_2 \prec \cdots \prec c_K,
\]
where $\prec$ denotes a strict ordering relation.
Additionally, structural consistency requires that the predicted classes at neighbouring pixels differ by at most 1.

To incorporate the natural ordering of labels into the learning process, a central idea of ordinal segmentation is to weight prediction errors according to the ordinal distance between the predicted and true labels.
This strategy can be implemented through ordinal loss functions, generically denoted as $\mathcal{L}_{\text{ord}}(k, \hat{\mathbf{p}})$, where $k$ denotes the index of the true class and $\hat{\mathbf{p}}$ the predicted probability distribution. These loss functions penalize ordinal inconsistencies by assigning higher costs to predictions that violate the natural ordering of the classes. In this way, errors between adjacent classes incur smaller penalties, while confusions between ordinally distant classes are progressively more penalized, explicitly reflecting the ordinal structure of the labels~\cite{gutierrez2016ordinal}.

The ordinal loss can be combined with the standard cross-entropy, resulting in a hybrid loss function that
preserves the discriminative capability of categorical classification while incorporating ordinal sensitivity.
The total loss function is therefore defined as
\begin{equation}
\mathcal{L} = \mathcal{L}_{CE} + \lambda \, \mathcal{L}_{\text{ord}},
\label{eq:loss}
\end{equation}
where \( \lambda \geq 0 \) is a hyperparameter that controls the trade-off between the traditional categorical
penalty and the order-sensitive penalty. This formulation allows ordinal knowledge to be integrated into the
training of deep neural networks without the need for significant architectural modifications.

\subsection{Ordinal Loss Functions}
\label{subsec:ordinal_losses}

As discussed previously, the loss function should go beyond simply maximizing the probability of the correct class for each pixel. It should incorporate the ordinal structure of the labels by penalizing classification errors asymmetrically. Confusions between ordinally adjacent classes should receive milder penalties, whereas errors involving ordinally distant classes should be progressively more penalized. A natural way to impose this asymmetry is to explicitly model the desired behavior of the predicted probability distributions, as presented next.

\subsubsection{Quasi-Unimodal Loss}
\label{subsubsec:quasi_unimodal_loss}

In ordinal problems, it is common to encourage the predicted probability distribution for each instance (or pixel, in the case of segmentation) to follow a \emph{unimodal} structure. For a pixel located at position $(i,j)$ in image $n$ with ground-truth label $y_{n,i,j} = k$, let the predicted probability vector 
\[
\hat{\mathbf{p}}_{n,i,j} = 
\left(
\hat{\mathbf{p}}_{n,i,j,1}, 
\hat{\mathbf{p}}_{n,i,j,2}, 
\dots, 
\hat{\mathbf{p}}_{n,i,j,K}
\right),
\]
represent the predicted distribution over the $K$ ordinal classes. Under \emph{strict unimodality}, the vector $\hat{\mathbf{p}}_{n,i,j}$ is expected to satisfy the monotonicity pattern

\begin{equation}
\hat{\mathbf{p}}_{n,i,j,1} 
\leq 
\cdots 
\leq 
\hat{\mathbf{p}}_{n,i,j,k} 
\geq 
\cdots 
\geq 
\hat{\mathbf{p}}_{n,i,j,K},
\end{equation}
meaning that the probabilities increase up to the true class $k$ and decrease afterwards, forming a single peak centered on the target class. This behavior is assumed to reflect the ordinal structure of the problem.

However, strict unimodality can be too restrictive in practical scenarios, where small local fluctuations are acceptable. While unimodality is a sufficient condition, it is not a necessary condition to achieve ordinal consistency.
To address this, the \emph{Quasi-Unimodal Loss} (QUL), proposed in~\cite{qul22}, relaxes the requirement of strict unimodality. 
In the \emph{quasi-unimodal} case, the vector $\hat{\mathbf{p}}_{n,i,j}$ does not need to increase systematically toward the peak and decrease afterwards
A probability distribution is quasi unimodal if the probability of left neighbour of the mode is higher than all the probabilities to the left (the left neighbour of the mode dominates all probabilities further to the left) and if the probability of right neighbour of the mode is higher than all the probabilities to the right (the right neighbour of the mode dominates all probabilities further to the right).
The resulting loss, averaged over all spatial locations, is defined as

\begin{equation}
\label{eq:qul}
\begin{aligned}
\mathcal{L}_{\text{QUL}}(\mathcal{D}_{tr}; \mathbf{w})
=
\frac{1}{N_{tr} H W}
\sum_{n=1}^{N_{tr}}
\sum_{i=1}^{H}
\sum_{j=1}^{W}
\Bigg[
&
\mathrm{ReLU}\!\left(\delta + \hat{y}_{n,i,j,k^{*}-1} - \hat{y}_{n,i,j, k^{*}}\right)
\\[4pt]
&+
\mathrm{ReLU}\!\left(\delta + \hat{y}_{n,i,j,k^{*}+1} - \hat{y}_{n,i,j,k^{*}}\right)
\\[4pt]
&+
\lambda
\sum_{(k,k') \in S^{A}(k^{*})}
\mathrm{ReLU}\!\left(\delta + \hat{y}_{n,i,j,k} - \hat{y}_{n,i,j,k'}\right)
\\[4pt]
&+
\lambda
\sum_{(k,k') \in S^{D}(k^{*})}
\mathrm{ReLU}\!\left(\delta + \hat{y}_{n,i,j,k} - \hat{y}_{n,i,j,k'}\right)
\Bigg].
\end{aligned}
\end{equation}

In Equation~\eqref{eq:qul}, the loss is computed over the training set $\mathcal{D}_{tr}$. For each pixel $(i,j)$ in image $n$, $k^{*}$ denotes the index of the ground-truth class and $\hat{y}_{n,i,j,k}$ the predicted probability for class k. The first two terms enforce local unimodality around the peak by comparing the predicted value of the peak class with those of its immediate neighbors $k^{*}-1$ and $k^{*}+1$. The remaining terms promote that the left neighbour of the mode dominates all probabilities further to the left and that the right neighbour of the mode dominates all probabilities further to the right.

The function $\mathrm{ReLU}(x) = \max(0,x)$ introduces a margin-based penalty controlled by $\delta > 0$, allowing limited local deviations without incurring loss. The parameter $\lambda > 0$ balances the contribution of the global ordinal constraints relative to the local neighborhood constraints. The sets are formally defined as
\begin{equation}
\label{eq:neighboring_sets}
\begin{aligned}
S^{A}(k^{*})
&= \{(k, k^{*}-1) \mid 1 \leq k \leq k^{*}-2\}, \\
S^{D}(k^{*})
&= \{(k^{*}+1, k) \mid k^{*}+2 \leq k \leq K\}.
\end{aligned}
\end{equation}

\subsubsection{Expectation Mean Squared Error}
\label{subsubsec:exp_mse_loss}
The \textit{Expectation Mean Squared Error} (\emph{EXP\_MSE}) loss, proposed in~\cite{expmse2026}, is designed to handle ordinal labels by explicitly modeling the expectation and variance of the predicted ordinal distribution. For each pixel $(i,j)$ of image $n$, the expected predicted ordinal label is defined as

\[
\mathbb{E}\!\left[ Y_{n,i,j} ; \textbf{x}_n ; \mathbf{w} \right]
=
\sum_{k=1}^{K} k \, \hat{\mathbf{p}}_{n,i,j, k},
\]
where $Y_{n,i,j}$ denotes the discrete ordinal random variable associated with the predicted class at that pixel. The corresponding variance is given by

\[
\mathrm{Var}\!\left[ Y_{n,i,j} ; \textbf{x}_n ; \mathbf{w} \right]
=
\sum_{k=1}^{K}
\hat{\mathbf{p}}_{n,i,j,k}
\left(
k -
\mathbb{E}\!\left[ Y_{n,i,j} ; \textbf{x}_n ; \mathbf{w} \right]
\right)^2.
\]

The expectation represents the predicted ordinal value obtained by weighting each class index by its probability, while the variance quantifies the dispersion of the predicted ordinal distribution.

Based on these quantities, the EXP\_MSE loss of $N_{tr}$ samples from $\mathcal{D}_{tr}$ is defined as

\begin{equation}
\mathcal{L}_{\text{EXP\_MSE}}(\mathcal{D}_{tr}; \mathbf{w})
=
\frac{1}{N_{tr} H W}
\sum_{n=1}^{N_{tr}}
\sum_{i=1}^{H}
\sum_{j=1}^{W}
\left(
\left|
\mathbb{E}\!\left[ Y_{n,i,j} ; \textbf{x}_n ; \mathbf{w} \right]
-
y_{n,i,j}
\right|^2
+
\lambda \,
\mathrm{Var}\!\left[ Y_{n,i,j} ; \textbf{x}_n ; \mathbf{w} \right]
\right),
\label{eq:expmse}
\end{equation}
where $\lambda > 0$ is a hyperparameter controlling the contribution of the variance regularization term.

The training procedure consists of solving the optimization problem

\[
\mathbf{w}^*
=
\arg\min_{\mathbf{w} \in \mathbb{R}^d}
\mathcal{L}_{\text{EXP\_MSE}}(\mathcal{D}_{tr}; \mathbf{w}).
\]

This loss exhibits desirable properties for ordinal segmentation tasks. In particular, predictions whose expected ordinal value is close to the ground-truth label and whose predicted distribution has low variance yield smaller loss values. Conversely, predictions that are ordinally distant from the true label or highly dispersed are penalized more strongly. Consequently, EXP\_MSE encourages both ordinal accuracy and confidence while preserving the intrinsic ordered structure of the labels.

\subsubsection{Ordinal Representation Consistency Loss}
\label{subsec:02}

To encourage ordinal consistency at the pixel level, Cruz \emph{et al.}  \cite{Cruz_2025}   extend concepts originally explored in ordinal classification to the semantic segmentation setting. In particular, previous works have proposed loss functions that explicitly enforce the ordered structure of labels during training~\cite{belharbi2019ordinal}. Building on these ideas, Cruz \emph{et al.} adapt ordinal constraints to pixel-wise predictions, encouraging the model to produce outputs that respect the ordinal relationships between classes in semantic segmentation tasks.

Within this framework, the ordinal representation consistency loss, referred to as O2, penalizes pixel-wise predictions that violate the expected ordering among class probabilities, explicitly encoding ordinal relationships during training. The O2 loss of $N_{tr}$ samples from $\mathcal{D}_{tr}$ is defined as

\begin{equation}
\label{eq:o2}
\begin{aligned}
\mathcal{L}_{\text{O2}}(\mathcal{D}_{tr};\mathbf{w})
=
\frac{1}{N_{tr} H W}
\sum_{n=1}^{N_{tr}}
\sum_{i=1}^{H}
\sum_{j=1}^{W}
\Bigg[
&
\sum_{(k,k') \in S_{N_B}^{-}(k^{*})}
\mathrm{ReLU}\!\left(
\delta + \hat{y}_{n,i,j,k'} - \hat{y}_{n,i,j,k}
\right)
\\[4pt]
&+
\sum_{(k,k') \in S_{N_B}^{+}(k^{*})}
\mathrm{ReLU}\!\left(
\delta + \hat{y}_{n,i,j,k'} - \hat{y}_{n,i,j,k}
\right)
\Bigg],
\end{aligned}
\end{equation}
where $S_{N_B}^{-}(k^{*})$ and $S_{N_B}^{+}(k^{*})$ denote the sets of neighboring ordinal class pairs immediately below and above the ground-truth class $k^{*}$, respectively (see Equation (\ref{eq:neighboring_sets})). The margin parameter $\delta > 0$ controls the minimum separation enforced between the predicted probabilities of ordered classes, and the $\mathrm{ReLU}$ operator ensures that only violations of the desired ordinal ordering contribute positively to the loss.

The sets are formally defined as
\begin{equation}
\label{eq:neighboring_sets}
\begin{aligned}
S_{N_B}^{-}(k^{*})
&= \{(k-1, k) \mid 2 \leq k \leq k^{*}\}, \\
S_{N_B}^{+}(k^{*})
&= \{(k, k+1) \mid k^{*}+1 \leq k \leq K\}.
\end{aligned}
\end{equation}

The O2 loss can be interpreted as a local ordinal regularization mechanism. In contrast, the QUL loss introduces a global quasi-unimodal constraint over the ordered label space. While O2 focuses on enforcing local ordinal relationships between neighboring classes, QUL regulates the overall shape of the predicted distribution, discouraging oscillations and secondary peaks while still allowing small local deviations from strict unimodality.

\subsubsection{Contact Surface Loss with Neighboring Pixels}
\label{subsubsec:csnp}

The contact surface loss based on neighboring pixels, proposed in~\cite{Cruz_2025}, aims to enforce local spatial ordinal consistency by penalizing situations in which spatially adjacent pixels are assigned to classes that are not ordinally adjacent.
In this way, abrupt transitions between labels that are distant in the ordinal space are discouraged, promoting smoother and more coherent ordinal boundaries.

For two spatially neighboring pixels $(i,j)$ and $(i',j')$ in image $n$, let $\hat{\mathbf{p}}_{n,i,j}, \hat{\mathbf{p}}_{n,i',j'} \in \mathbb{R}^K$ denote their respective predicted probability distributions over the $K$ ordinal classes. The pairwise penalty between these neighboring pixels is defined through the bilinear form

\begin{equation}
\mathcal{L}_{\text{CSNP}}^{(n,i,j,i',j')}(\mathbf{w})
=
\hat{\mathbf{p}}_{n,i,j}^{\top}
\mathbf{C}
\,
\hat{\mathbf{p}}_{n,i',j'},
\label{eq:csnp}
\end{equation}
where $\mathbf{C} \in \mathbb{R}^{K \times K}$ is a cost matrix encoding the ordinal distance between classes. Its entries are defined as

\begin{equation}
\label{eq:cost-matrix}
C_{r,s}
=
\max\bigl(0, |r - s| - 1\bigr) = \text{ReLU}(|r - s| - 1),
\end{equation}
where $r,s \in \{1,\ldots,K\}$ denote ordinal class indices.
The value $C_{r,s}$ assigns zero cost to identical or immediately adjacent classes, and increases proportionally with the ordinal distance between non-adjacent classes.

The overall contact surface loss of $N_{tr}$ images sampled from $\mathcal{D}_{tr}$ is obtained by summing the pairwise penalties over all neighboring pixel pairs $\mathcal{N}$:

\[
\mathcal{L}_{\text{CSNP}}(\mathcal{D}_{tr};\mathbf{w})
=
\frac{1}{|\mathcal{N}|}
\sum_{n=1}^{N_{tr}}
\sum_{((i,j),(i',j')) \in \mathcal{N}}
\hat{\mathbf{p}}_{n,i,j}^{\top}
\mathbf{C}
\hat{\mathbf{p}}_{n,i',j'}.
\]

Consequently, this formulation imposes no penalty when neighboring pixels are assigned to equal or ordinally adjacent classes, while progressively penalizing interactions between ordinally distant classes. This mechanism effectively discourages inconsistent predictions in spatially adjacent regions and promotes ordinally coherent segmentation boundaries.

\subsubsection{Contact Surface Loss with Distance Transform}
\label{subsubsec:csdt}
An alternative approach to enforcing spatial ordinal consistency, proposed in~\cite{Cruz_2025}, is based on the use of the Distance Transform (DT), which provides, for each spatial location, the distance to the nearest high-confidence pixel belonging to a given class.

For a fixed class $k \in \{1,\dots,K\}$ and image $n$, define the predicted probability map
\[
\hat{\mathbf{p}}_{n,k} : \Omega \subset \mathbb{Z}^2 \to [0,1]^{H\times W},
\]
where $\Omega = \{1,\dots,H\} \times \{1,\dots,W\}$ denotes the spatial grid.

The distance transform associated with class $k$ is defined, for each spatial location $\mathbf{x} \in \Omega$, as

\begin{equation}
\mathrm{DT}_{n,k}(\mathbf{x})
=
\min_{\mathbf{y} \in \Omega \, : \, \hat{\mathbf{p}}_{n,k}(\mathbf{y}) \geq \delta}
d(\mathbf{x},\mathbf{y}),
\label{eq:dt}
\end{equation}
where $\delta \in (0,1)$ is a confidence threshold selecting pixels strongly associated with class $k$, and $d(\cdot,\cdot)$ denotes a spatial distance metric, typically the Euclidean distance.

To avoid excessive penalization and training instabilities, the distance transform is bounded by a maximum value $\gamma > 0$, yielding the saturated version

\[
\mathrm{DT}_{n,k}^{\gamma}(\mathbf{x})
=
\min\bigl(\mathrm{DT}_{n,k}(\mathbf{x}), \gamma\bigr).
\]

The loss is formulated to encourage spatial separation between regions associated with ordinally distant classes. For this purpose, the probability map of one class is combined with the distance transform of another class.

Let
\begin{equation}
\label{eq:class-set}
\mathcal{S}
=
\{(k_1,k_2) \in \{1,\dots,K\};  \; |k_1 - k_2| > 1\},
\end{equation}
denotes the set of ordinally non-adjacent class pairs, and let $\mathbf{C}$ be the cost matrix defined in Equation~\eqref{eq:cost-matrix}, with entries $C_{k_1,k_2}$ encoding the ordinal distance between classes. The Contact Surface Distance Transform loss  of $N_{tr}$ samples from $\mathcal{D}_{tr}$ is then defined as

\begin{equation}
\mathcal{L}_{\text{CSDT}}(\mathcal{D}_{tr};\mathbf{w})
=
- \frac{1}{N_{tr} |\Omega|}
\sum_{n=1}^{N_{tr}}
\sum_{(k_1,k_2) \in \mathcal{S}}
C_{k_1,k_2}
\sum_{\mathbf{x} \in \Omega}
\left(
\hat{\mathbf{p}}_{n,k_1}(\mathbf{x}) \,
\mathrm{DT}_{n,k_2}^{\gamma}(\mathbf{x})
+
\hat{\mathbf{p}}_{n,k_2}(\mathbf{x}) \,
\mathrm{DT}_{n,k_1}^{\gamma}(\mathbf{x})
\right),
\label{eq:csdt}
\end{equation}
where $|\Omega| = H W$.

This formulation penalizes configurations in which pixels assigned to a high probability of an ordinal class are spatially close to regions strongly associated with ordinally distant classes. The saturation parameter $\gamma$ ensures that only spatially close but ordinally inconsistent regions contribute significantly to the loss, thereby promoting smoother and more coherent transitions in the segmentation maps while respecting the ordinal structure of the label space.

\subsubsection{Contact Surface Loss with Signed Distance Function Loss}
\label{subsubsec:cssdf}

The Contact Surface Loss with Signed Distance Function (CSSDF) is proposed here as a direct extension of the Contact Surface Loss with Distance Transform (CSDT) introduced in Equation \cite{Cruz_2025}. In this formulation, the unsigned Distance Transform is used to construct a Signed Distance Function (SDF), allowing the loss to explicitly encode both interior and exterior object information. This modification enriches the spatial modeling of transitions between ordinally distant classes and leads to more accurate and well-defined object boundaries.

Based on the distance transform $\mathrm{DT}_{n,k}$ defined in
Equation (~\ref{eq:dt}), the predicted signed distance function for class $k$ is defined as

\begin{equation}
\hat{\phi}_{n,k}(\mathbf{x})
=
\begin{cases}
\;\;\; \mathrm{DT}_{n,k}(\mathbf{x}), 
& \hat{\mathbf{p}}_{n,k}(\mathbf{x}) \geq \delta, \\
- \mathrm{DT}_{n,k}(\mathbf{x}), 
& \hat{\mathbf{p}}_{n,k}(\mathbf{x}) < \delta,
\end{cases}
\end{equation}
where $\delta \in (0,1)$ is the confidence threshold used to define
high-probability regions.

Similarly, let $\mathbf{p}_{n,k} : \Omega \to \{0,1\}$ denote the
ground-truth one-hot map for class $k$. The corresponding ground-truth signed distance function is defined as

\begin{equation}
\phi_{n,k}(\mathbf{x})
=
\begin{cases}
\;\;\; \mathrm{DT}^{gt}_{n,k}(\mathbf{x}), 
& \mathbf{p}_{n,k}(\mathbf{x}) = 1, \\
- \mathrm{DT}^{gt}_{n,k}(\mathbf{x}), 
& \mathbf{p}_{n,k}(\mathbf{x}) = 0,
\end{cases}
\end{equation}
where $\mathrm{DT}^{gt}_{n,k}$ denotes the distance transform computed from the ground-truth mask of class $k$.

The CSSDF loss of $N_{tr}$ samples from $\mathcal{D}_{tr}$ is defined as

\begin{equation}
\mathcal{L}_{\text{CSSDF}}(\mathcal{D}_{tr};\mathbf{w})
=
\frac{1}{N_{tr}|\Omega|}
\sum_{n=1}^{N_{tr}}
\sum_{\mathbf{x} \in \Omega}
\sum_{(k_1,k_2) \in \mathcal{S}}
C_{k_1,k_2}
\left(
\alpha_{n,k_1}(\mathbf{x})
\bigl|
\phi_{n,k_2}^{\hat{\gamma}}(\mathbf{x})
-
\hat{\phi}_{n,k_2}^{\hat{\gamma}}(\mathbf{x})
\bigr|^p
+
\alpha_{n,k_2}(\mathbf{x})
\bigl|
\phi_{n,k_1}^{\hat{\gamma}}(\mathbf{x})
-
\hat{\phi}_{n,k_1}^{\hat{\gamma}}(\mathbf{x})
\bigr|^p
\right),
\label{eq:cssdf}
\end{equation}
where $p \geq 1$ controls the penalty strength and

\[
\alpha_{n,k}(\mathbf{x})
=
\exp\bigl(-\gamma\, |\hat{\phi}_{n,k}(\mathbf{x})|\bigr),
\quad \gamma > 0,
\]
is a spatial weighting factor that emphasizes pixels close to the class boundaries.

As in the CSDT formulation, the signed distance functions are bounded to avoid excessive penalization:

\begin{equation}
\phi_{n,k}^{\hat{\gamma}}(\mathbf{x})
=
\mathrm{sign}\bigl(\phi_{n,k}(\mathbf{x})\bigr)
\min\bigl(|\phi_{n,k}(\mathbf{x})|, \hat{\gamma}\bigr),
\end{equation}

and analogously for $\hat{\phi}_{n,k}^{\hat{\gamma}}$.

This formulation preserves the cumulative structure of CSDT while leveraging signed distance information, enabling explicit distinction between interior and exterior object regions. Consequently, CSSDF promotes improved boundary accuracy and enhanced geometric coherence in transitions between ordinally distant classes.

\subsection{Evaluation Metrics}
\label{subsec:evaluation_metrics}

The performance of the proposed methods is evaluated using a set of metrics that go beyond pixel-wise accuracy, incorporating aspects of ordinal coherence and structural consistency of the predicted segmentations, as detailed below.

\subsubsection{Unimodal Pixels Metric}
\label{subsubsec:unimodal_pixels}
For each pixel \( \mathbf{x} \in \Omega \), where:
\begin{equation}
\label{eq:omega}
\Omega =  \{1,\dots,H\} \times \{1,\dots,W\}, 
\end{equation}
an indicator function of pixel-wise unimodality is defined as
\begin{equation}
\mathbb{1}_{\mathrm{UP}}(\mathbf{x}) =
\begin{cases}
1, & \text{if } \hat{\mathbf{p}}(\mathbf{x}) \text{ is unimodal}, \\
0, & \text{otherwise}.
\end{cases}
\end{equation}

The \emph{Unimodal Pixels (UP)} metric~\cite{Cardoso_2025} is defined as the fraction of pixels whose predicted probability distribution satisfies the unimodality property:
\begin{equation}
\mathrm{UP} =
\frac{1}{|\Omega|}
\sum_{\mathbf{x} \in \Omega}
\mathbb{1}_{\mathrm{UP}}(\mathbf{x}).
\label{eq:UPmetric}
\end{equation}

This metric directly evaluates the ordinal consistency of the probabilistic outputs produced by the
model, independently of the final class decision, making it particularly suitable for ordinal semantic
segmentation tasks.

\subsubsection{Contact Surface Metric}
\label{subsubsec:contact_surface_metric}
To assess the spatial structural consistency of ordinal segmentations, the \emph{Contact Surface} (CS) metric, proposed in~\cite{Cruz_2025}, is employed. This metric quantifies the frequency of ordinally invalid transitions between spatially adjacent pixels, reflecting the geometric regularity of class boundaries. The corresponding predicted label map is defined pointwise as
\begin{equation}
\hat{y}_n(i,j)
=
\argmax_{k \in \{1,\dots,K\}}
\hat{\mathbf{p}}_{n,k}(i,j),
\qquad (i,j) \in \Omega.
\end{equation}
where $\Omega$ is defined in Equation (~\ref{eq:omega}).

In the context of ordinal segmentation, transitions between classes
with ordinal distance equal to $1$ are considered valid. In contrast, direct transitions between classes with ordinal distance greater than $1$ correspond to invalid ordinal jumps.

For each pixel $(i,j) \in \Omega$, the horizontal and vertical ordinal variations are defined as

\begin{equation}
\mathrm{CS}_{dx}(\hat{y}_n)_{i,j}
=
\bigl| \hat{y}_n(i,j) - \hat{y}_n(i,j+1) \bigr|,
\quad
(i,j) \in  \Omega
\end{equation}

\begin{equation}
\mathrm{CS}_{dy}(\hat{y}_n)_{i,j}
=
\bigl| \hat{y}_n(i,j) - \hat{y}_n(i+1,j) \bigr|,
\quad
(i,j) \in \Omega.
\end{equation}

The CS metric for image $x_n$ is then defined as

\begin{equation}
\mathrm{CS}(x_n)
=
\frac{1}{2}
\left(
\frac{
\sum_{i=1}^{H} \sum_{j=1}^{W-1}
\mathbb{1}\!\left( \mathrm{CS}_{dx}(\hat{y}_n)_{i,j} \geq 2 \right)
}{
\sum_{i=1}^{H} \sum_{j=1}^{W-1}
\mathbb{1}\!\left( \mathrm{CS}_{dx}(\hat{y}_n)_{i,j} \geq 1 \right)
+
\varepsilon
}
+
\frac{
\sum_{i=1}^{H-1} \sum_{j=1}^{W}
\mathbb{1}\!\left( \mathrm{CS}_{dy}(\hat{y}_n)_{i,j} \geq 2 \right)
}{
\sum_{i=1}^{H-1} \sum_{j=1}^{W}
\mathbb{1}\!\left( \mathrm{CS}_{dy}(\hat{y}_n)_{i,j} \geq 1 \right)
+
\varepsilon
}
\right),
\label{eq:CSmetric}
\end{equation}
where $\mathbb{1}(\cdot)$ denotes the indicator function and
$\varepsilon > 0$ is a small constant introduced to avoid division by zero. Lower values of $\mathrm{CS}(x_n)$ indicate higher spatial ordinal
coherence, as they correspond to a reduced proportion of abrupt
transitions between ordinally distant classes. The final dataset-level metric is obtained by averaging $\mathrm{CS}(x_n)$ over all test images.

\subsection{Statistical Performance Evaluation}
\label{subsec:statistical-performace}

Through the cross-validation procedure, $K$ performance estimates are obtained, from which the mean $\mu$ and the standard deviation $\sigma$ of the observed accuracies are computed. These values allow the construction of statistical intervals of the form
$I = [\mu - \sigma, \mu + \sigma]$, which provide a measure of the variability associated with the neural network performance.

Given a pair of intervals
$I_1 = [\mu_1 - \sigma_1, \mu_1 + \sigma_1]$ and
$I_2 = [\mu_2 - \sigma_2, \mu_2 + \sigma_2]$, it is necessary to establish a criterion to identify which interval demonstrates superior performance.  In this work, this decision is based on the methodology proposed in~\cite{de2021pairwise}, according to which the interval $I_1$ is considered worse than the interval$I_2$ when one of the following conditions is satisfied:

\begin{flalign}
\label{eq:liliane}
&I_{1} \text{ is considered inferior to } I_{2} \text{ if:} & \\ \nonumber
&(a)\; \mu_{1} < \mu_{2} \text{ and } I_{1} \cap I_{2} = \emptyset, & \\ \nonumber
&(b)\; \mu_{1} < \mu_{2} \text{ and } I_{1} \cap I_{2} = I_{1}, & \\ \nonumber
&(c)\; \mu_{1} < \mu_{2} \text{ and } (\mu_{1} + \sigma_{1}) < \mu_{2}, & \\ \nonumber
&(d)\; \mu_{1} < \mu_{2} \text{ and } (\mu_{1} + \sigma_{1}) < \mu_{2} + \sigma_{2}, & \\ \nonumber
&(e)\; \mu_{1} < \mu_{2} \text{ and }
\varrho = \frac{|(\mu_{1} + \sigma_{1}) - (\mu_{2} - \sigma_{2})|}{2\sigma_{1}}
\text{ assumes small values.}
\end{flalign}

\section{Experimental Evaluation}
\label{sec:Experimental-evaluation}

In order to perform a fair comparison between the results obtained using the loss functions presented in Sections~\ref{subsubsec:quasi_unimodal_loss},~\ref{subsubsec:exp_mse_loss}, and~\ref{subsubsec:cssdf} and the methods proposed in~\cite{Cruz_2025}, exactly the same datasets employed in that study were used. These datasets consist of medical images acquired from different modalities and clinical applications, covering distinct ordinal segmentation scenarios.

\begin{figure}[H]
    \centering
    \includegraphics[width=0.8\linewidth]{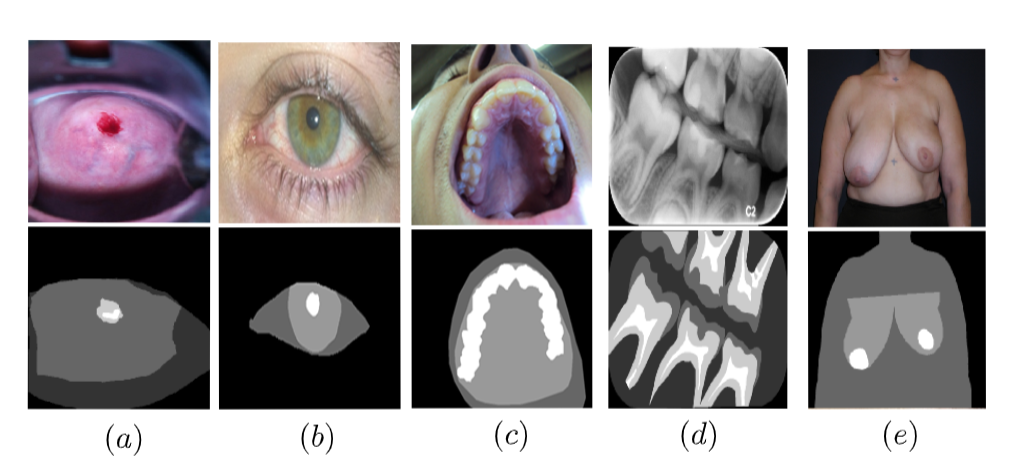}
    \caption{Sample images and their corresponding ordinal masks from each dataset. The datasets are organized as follows: (a) Breast Cervix-MobileODT; (b) Mobbio; (c)Teeth-UCV; (d) Teeth-ISBI; and (e) Breast Aesthetics.}
    \label{fig:dataset}
\end{figure}

Specifically, the following datasets were considered: Breast Aesthetics~\cite{CARDOSO2007115}, Cervix-MobileODT~\cite{intel-mobileodt-cervical-cancer-screening}, Mobbio~\cite{mobio}, Teeth-ISBI~\cite{WANG201663}, and Teeth-UCV~\cite{teethucv}. Figure~\ref{fig:dataset} presents representative samples of the images and their corresponding ordinal masks.

For the segmentation task, the U-Net architecture was adopted, trained using normalized data and data augmentation techniques, strictly following the same experimental settings described in~\cite{Cruz_2025}. Training was performed using the Adam optimizer, a learning rate of $10^{-4}$, and a batch size of $16$, employing early stopping with a patience of $15$ epochs. The maximum number of epochs was set to $200$.

The data were split into training and test sets using an $80\%$–$20\%$ proportion. In addition, 5-fold cross-validation was applied, and in each fold, the model with the best performance on the validation set was selected. The evaluation metrics adopted are also the same as those used in~\cite{Cruz_2025}, ensuring consistency in the comparison of results.

Performance evaluation was considered three main metrics: (i) the Dice coefficient~\cite{mueller2022towards}, (ii) the contact surface (Equation (\ref{eq:CSmetric})), and (iii) the percentage of unimodal pixels (Equation (\ref{eq:UPmetric})). Cross-entropy was adopted as the reference loss. However, the optimization objective corresponds to the combined formulation described in Equation (\ref{eq:loss}), with $\lambda \in {0.1, 1, 10, \ldots, 10^4}$. The ordinal loss functions investigated include QUL (Equation (\ref{eq:qul})), EXP\_MSE (Equation (~\ref{eq:expmse})), and CSSDF (Equation ( ~\ref{eq:cssdf})), as well as combinations of unimodal loss functions with the spatial loss, namely QUL + CSSDF and EXP\_MSE + CSSDF.

The hyperparameters adopted for each loss function are described as follows:

\begin{itemize}
    \item Ordinal loss function QUL: $\delta \in [0.05, 0.7]$ and $\lambda \in \{0.1, 1, 10, \ldots, 10^4\}$;
    \item Ordinal loss function EXP\_MSE: $\lambda \in \{0.1, 1, 10, \ldots, 10^4\}$;
    \item Ordinal loss function CSSDF: $\gamma \in [0.05, 1.0]$, $\delta = 0.05$, and $p \in \{1, 2\}$.
\end{itemize}

\section{Results}
\label{sec:results}

The results presented in this section highlight the performance of the U-Net architecture in the task of ordinal semantic segmentation of medical images, considering the use of different unimodal and spatial ordinal loss functions. The quantitative results are reported in Tables~\ref{tab:loss_comparison_CS}-\ref{tab:structured_loss_comparison_UP}, where the best performance results in each row are highlighted in \textcolor{blue}{blue}.  The comparison between the intervals defined by the mean and the standard deviation was performed based on the criterion presented in Equation~\eqref{eq:liliane}. The evaluation was conducted using 5-fold cross-validation, from which the mean accuracy and standard deviation were computed and analyzed, as described in Section~\ref{sec:Experimental-evaluation}. The datasets used in this study are presented in Figure~\ref{fig:dataset}.

\begin{table}[H]
    \centering
    \caption{Overall results of the contact surface metric (\%), a structural metric that evaluates consistency between adjacent pixels. Lower values indicate better performance.}
    \label{tab:loss_comparison_CS}
    \scalebox{0.75}{
    \begin{tabular}{lcccc}
        \toprule
        Dataset 
        & CE 
        & CE + QUL 
        & CE + EXP\_MSE 
        & CE + CSSDF \\
        \midrule
        Breast Aesthetics & $0.2 \pm 0.2$ & \textcolor{blue}{$0.1 \pm 0.0$} & \textcolor{blue}{$0.1 \pm 0.0$} & $0.1 \pm 0.1$ \\
        Cervix-MobileODT & $14.5 \pm 3.3$ & $1.2 \pm 0.5$ & \textcolor{blue}{$1.0 \pm 0.6$} & $13.2 \pm 1.5$ \\
        Mobbio & $12.3 \pm 0.5$ & \textcolor{blue}{$3.8 \pm 0.1$} & $3.8 \pm 0.5$ & $12.1 \pm 0.3$ \\
        Teeth-ISBI & $30.0 \pm 3.9$ & \textcolor{blue}{$26.2 \pm 2.5$}& $26.8 \pm 2.1$ & $27.2 \pm 1.5$ \\
        Teeth-UCV & $7.0 \pm 1.4$ & \textcolor{blue}{$3.1 \pm 0.6$} & $5.9 \pm 0.9$ & $5.6 \pm 0.8$ \\
        \bottomrule
    \end{tabular}}
\end{table}

Table~\ref{tab:loss_comparison_CS} reports the overall results obtained with the contact surface metric (Equation (~\ref{eq:CSmetric})), which evaluates structural consistency between adjacent pixels and favors lower values. Across all datasets, the incorporation of unimodal ordinal losses (QUL and EXP\_MSE) consistently reduces the contact surface error when compared to the standard cross-entropy loss.

In particular, the Breast Aesthetics dataset achieves the lowest values with both CE+QUL and CE+EXP\_MSE. For Cervix-MobileODT, CE+EXP\_MSE yields the best performance, substantially reducing structural inconsistencies relative to CE. In the Teeth datasets, CE+QUL provides the lowest contact surface values, highlighting its effectiveness in preserving ordinal transitions in more complex anatomical structures. In contrast, the structured loss CE+CSSDF shows limited gains in this metric, especially for Cervix-MobileODT, Mobbio, and Teeth-ISBI.

\begin{table}[H]
    \centering
    \caption{Overall results for the unimodal pixels metric (\%), a representation metric that evaluates ordinal consistency at the pixel level. Higher values indicate better performance.}
    \label{tab:loss_comparison_unimodal}
    \scalebox{0.75}{
    \begin{tabular}{lcccc}
        \toprule
        Dataset 
        & CE 
        & CE + QUL 
        & CE + EXP\_MSE 
        & CE + CSSDF \\
        \midrule
        Breast Aesthetics & $6.2 \pm 0.7$ & \textcolor{blue}{$7.3 \pm 0.6$} & $6.5 \pm 2.0$ & $7.0 \pm 1.2$ \\
        Cervix-MobileODT & $1.0 \pm 0.2$ & $1.0 \pm 0.2$ & \textcolor{blue}{$1.6 \pm 0.3$} & $0.8 \pm 0.2$ \\
        Mobbio & $0.8 \pm 0.1$ & $0.9 \pm 0.2$ & \textcolor{blue}{$0.9 \pm 0.3$} & $0.7 \pm 0.3$ \\
        Teeth-ISBI & $9.6 \pm 3.4$ & $12.3 \pm 4.2$ & $12.1 \pm 3.8$ & \textcolor{blue}{$13.7 \pm 3.5$} \\
        Teeth-UCV & $18.0 \pm 1.8$ & $25.5 \pm 7.2$ & \textcolor{blue}{$25.8 \pm 6.9$} & $20.1 \pm 5.0$ \\
        \bottomrule
    \end{tabular}}
\end{table}

Table~\ref{tab:loss_comparison_unimodal} presents the results for the unimodal pixels metric (Equation (~\ref{eq:UPmetric})), which measures ordinal consistency at the pixel level and favors higher values. The results show that unimodal losses generally improve pixel-wise ordinal coherence when compared to CE alone. For the Teeth-ISBI dataset, CE+CSSDF achieves the highest percentages of unimodal pixels. For the Teeth-UCV, Mobbio, and Cervix-MobileODT datasets, CE+EXP\_MSE yields the best performance. The CE+QUL combination achieves the best result only for the Breast Aesthetics dataset.

The segmentation accuracy in terms of the Dice coefficient is summarized in Table~\ref{tab:loss_comparison_dice}. Overall, the inclusion of ordinal loss functions leads to modest but consistent improvements over the baseline CE. Examining the values in the table, it can be seen that the highest Dice scores are achieved with CE+QUL for Breast Aesthetics, Teeth-ISBI, and Teeth-UCV, while CE+EXP\_MSE attains the highest values for Cervix-MobileODT and Mobbio, confirming that these loss combinations truly provide the best results for each dataset.

These results suggest that enforcing ordinal relationships during training does not compromise segmentation accuracy and can, in several cases, further stabilize segmentation performance. This occurs because ordinal losses encourage the model to respect the natural hierarchy of the classes, promoting structural coherence and spatial continuity, thereby improving qualitative segmentation behavior without degrading the Dice coefficient.

Comparisons between the unimodal loss functions proposed in this work, namely, QUL (Section \ref{subsubsec:quasi_unimodal_loss}) and EXP\_MSE (Section \ref{subsubsec:exp_mse_loss}), and the unimodal loss function previously reported in \cite{Cruz_2025}, and later reviewed in Section \ref{subsec:02}, are presented in Tables~\ref{tab:unimodal_loss_comparison_CS}–\ref{tab:unimodal_loss_comparison_DC}. Evaluation was performed using the Contact Surface metric (Section \ref{subsubsec:contact_surface_metric}), the percentage of unimodal pixels (Section \ref{subsubsec:unimodal_pixels}), and the Dice coefficient, respectively.

In terms of contact surface (Table~\ref{tab:unimodal_loss_comparison_CS}), CE combined with QUL consistently achieves the lowest values on the Breast Aesthetics dataset, outperforming both the baseline CE and the CE+O2 formulation. On the Mobbio dataset, both CE+QUL and CE+EXP\_MSE demonstrate superior performance compared to CE and CE+O2. Notably, CE+EXP\_MSE provides the best performance for Cervix-MobileODT, while CE+O2 yields the best results for Teeth-ISBI.

\begin{table}[H]
    \centering
    \caption{Overall results for the Dice coefficient metric (\%). Higher values indicate better performance.}
    \label{tab:loss_comparison_dice}
    \scalebox{0.75}{
    \begin{tabular}{lcccc}
        \toprule
        Dataset 
        & CE 
        & CE + QUL 
        & CE + EXP\_MSE 
        & CE + CSSDF \\
        \midrule
        Breast Aesthetics & $93.8 \pm 0.5$ & \textcolor{blue}{$94.4 \pm 0.4$} & $94.4 \pm 0.2$ & $93.7 \pm 0.8$ \\
        Cervix-MobileODT & $77.0 \pm 0.7$ & $77.2 \pm 0.2$ & \textcolor{blue}{$77.3 \pm 0.2$} & $76.8 \pm 0.3$ \\
        Mobbio & $93.8 \pm 0.1$ & $93.8 \pm 0.5$ & \textcolor{blue}{$94.0 \pm 0.1$} & $93.8 \pm 0.1$ \\
        Teeth-ISBI & $74.0 \pm 1.6$ & \textcolor{blue}{$75.6 \pm 0.3$} & $74.6 \pm 0.2$ & $75.3 \pm 1.5$ \\
        Teeth-UCV & $90.2 \pm 0.5$ & \textcolor{blue}{$90.8 \pm 0.4$} & $90.2 \pm 0.4$ & $90.1 \pm 0.1$ \\
        \bottomrule
    \end{tabular}}
\end{table}

\begin{table}[H]
    \centering
    \caption{Comparison of the unimodal loss functions proposed in this work with the CE+O2 (Section \ref{subsec:02}) loss presented in \cite{Cruz_2025}, using the contact surface metric ($\%$). Again, lower values indicate better performance.}
    \label{tab:unimodal_loss_comparison_CS}
    \scalebox{0.75}{
    \begin{tabular}{lcccc}
        \toprule
        Dataset 
        & CE 
        & CE + O2 
        & CE + QUL 
        & CE + EXP\_MSE \\
        \midrule
        Breast Aesthetics & $0.2 \pm 0.2$ & $0.2 \pm 0.2$ & \textcolor{blue}{$0.1 \pm 0.0$} & \textcolor{blue}{$0.1 \pm 0.0$} \\
        Cervix-MobileODT  & $14.5 \pm 3.3$ & $1.1 \pm 0.4$ & $1.2 \pm 0.5$ & \textcolor{blue}{$1.0 \pm 0.6$} \\
        Mobbio   & $ 12.3 \pm 0.5$ & $4.2 \pm 0.2$ & \textcolor{blue}{$3.8 \pm 0.1$} &$3.8 \pm 0.5$  \\
        Teeth-ISBI        & $30.0 \pm 3.9$ & \textcolor{blue}{$26.1 \pm 2.4$} & $26.2 \pm 2.5$ & $26.8 \pm 2.1$ \\
        Teeth-UCV         & $7.0 \pm 1.4$ & \textcolor{blue}{$3.0 \pm 0.7$} & $3.1 \pm 0.6$ &  $5.9 \pm 0.9$  \\
        \bottomrule
    \end{tabular}}
\end{table}

\begin{table}[H]
    \centering
    \caption{Comparison of the unimodal loss functions investigated in this work with the CE+02  (Section \ref{subsec:02}) loss presented in \cite{Cruz_2025}, using the percentage of unimodal pixels metric ($\%$). Again, higher values indicate better performance.}
    \label{tab:unimodal_loss_comparison_unimodalpixels}
    \scalebox{0.75}{
    \begin{tabular}{lcccc}
        \toprule
        Dataset 
        & CE 
        & CE + O2 
        & CE + QUL 
        & CE + EXP\_MSE \\
        \midrule
        Breast Aesthetics & $6.2 \pm 0.7$ & \textcolor{blue}{$8.1 \pm 2.6$} & $7.3 \pm 0.6$ & $6.5 \pm 2.0$  \\
        Cervix-MobileODT  & $1.0 \pm 0.2$ & \textcolor{blue}{$98.9 \pm 0.4$} & $1.0 \pm 0.2$ & $1.6 \pm 0.3$ \\
        Mobbio   & $ 0.8 \pm 0.1$ & \textcolor{blue}{$97.3 \pm 2.6$} & $0.9 \pm 0.2$ & $0.9 \pm 0.3$  \\
        Teeth-ISBI        & $9.6 \pm 3.4$ & \textcolor{blue}{$97.1 \pm 3.8$} & $12.3 \pm 4.2$ &  $12.1 \pm 3.8$\\
        Teeth-UCV         & $18.0 \pm 1.8$ & \textcolor{blue}{$96.8 \pm 1.0$} & $25.5 \pm 7.2$& $25.8 \pm 6.9$  \\
        \bottomrule
    \end{tabular}}
\end{table}

With respect to the unimodal pixels metric (Table~\ref{tab:unimodal_loss_comparison_unimodalpixels}), the CE+O2 combination exhibits significantly higher values across all datasets, reflecting its strong enforcement of pixel-level unimodality. However, these gains do not consistently translate into improvements in either the contact surface metric or the Dice coefficient, indicating that pixel-level ordinal coherence, when applied in isolation, is insufficient to ensure spatial consistency or enhance segmentation accuracy.

In contrast, the CE+QUL and CE+EXP\_MSE combinations exhibit a more balanced behavior across the different evaluation criteria. Although they yield lower unimodal pixel values compared to CE+O2, these losses produce lower contact surface errors and comparable or higher Dice coefficients, indicating improved structural consistency and overall segmentation quality. These findings suggest that an excessively strong enforcement of pixel-wise unimodality, as in CE+O2, may lead to overly local constraints that fail to adequately capture global spatial relationships. Conversely, QUL and EXP\_MSE promote ordinal consistency in a more balanced manner, preserving structural coherence while maintaining segmentation accuracy, resulting in more robust and reliable segmentations across different datasets.

\begin{table}[H]
    \centering
    \caption{Comparison of the unimodal loss functions investigated in this work with the CE+02 (Section \ref{subsec:02}) loss presented in \cite{Cruz_2025}, using the Dice coefficient metric ($\%$). Again, higher values indicate better performance.}
    \label{tab:unimodal_loss_comparison_DC}
    \scalebox{0.75}{
    \begin{tabular}{lcccc}
        \toprule
        Dataset 
        & CE 
        & CE + O2 
        & CE + QUL 
        & CE + EXP\_MSE \\
        \midrule
        Breast Aesthetics & $93.8 \pm 0.5$ & $94.2 \pm 0.6$ & \textcolor{blue}{$94.4 \pm 0.4$} & $94.4 \pm 0.2$ \\
        Cervix-MobileODT  & $77.0 \pm 0.7$ & $77.2 \pm 0.4$ & $77.2 \pm 0.2$ & \textcolor{blue}{$77.3 \pm 0.2$} \\
        Mobbio   & $ 93.8 \pm 0.1$ & \textcolor{blue}{$94.1 \pm 0.0$} & $93.8 \pm 0.5$ & $94.0 \pm 0.1$  \\
        Teeth-ISBI        & $74.0 \pm 1.6$ & $74.8 \pm 0.4$& \textcolor{blue}{$75.6 \pm 0.3$} & $74.6 \pm 0.2$ \\
        Teeth-UCV         & $90.2 \pm 0.5$ & \textcolor{blue}{$90.8 \pm 0.4$} & $90.8 \pm 0.4$ & $90.2 \pm 0.4$ \\
        \bottomrule
    \end{tabular}}
\end{table}

Table~\ref{tab:unimodal_loss_comparison_DC} shows that the highest Dice coefficients vary across the different loss functions: CE+QUL achieves the best results for Breast Aesthetics and Teeth-ISBI, while CE+EXP\_MSE attains the best performance for Cervix-MobileODT. This improvement in the Dice coefficient indicates higher segmentation accuracy, reflecting a stronger agreement between the predicted segmentation and the reference labels, which contributes to more reliable and consistent quantitative results.

\begin{table}[H]
    \centering
    \caption{Comparison of the structured loss functions investigated in this work with the CE+CSNP (Section \ref{subsubsec:contact_surface_metric}) and CE+CSDT (Section \ref{subsubsec:csdt}) losses presented in \cite{Cruz_2025}, using the contact surface metric (\%). Again, lower values indicate better performance.}
    \label{tab:structured_loss_comparison_CS}
    \scalebox{0.75}{
    \begin{tabular}{lcccc}
        \toprule
        Dataset 
        & CE 
        & CE + CSNP 
        & CE + CSDT
        & CE + CSSDF \\
        \midrule
        Breast Aesthetics & $0.2 \pm 0.2$ & \textcolor{blue}{$0.1 \pm 0.1$} & $0.2 \pm 0.2$ & \textcolor{blue}{$0.1 \pm 0.1$} \\
        Cervix-MobileODT  & $14.5 \pm 3.3$ & \textcolor{blue}{$10.8 \pm 2.7$} & $13.6 \pm 1.7$ & $13.2 \pm 1.5$ \\
        Mobbio            & $12.3 \pm 0.5$ & $12.3 \pm 0.4$ & $12.4 \pm 0.6$ & \textcolor{blue}{$12.1 \pm 0.3$} \\
        Teeth-ISBI        & $30.0 \pm 3.9$ & $28.4 \pm 1.4$ & \textcolor{blue}{$27.1 \pm 1.8$} & $27.2 \pm 1.5$\\
        Teeth-UCV         & $7.0 \pm 1.4$ & \textcolor{blue}{$2.3 \pm 1.0$} & $5.5 \pm 1.0$ & $5.6 \pm 0.8$ \\
        \bottomrule
    \end{tabular}}
\end{table}

\begin{table}[H]
    \centering
    \caption{Comparison of the structured loss functions investigated in this work with the CE+CSNP (Section \ref{subsubsec:contact_surface_metric}) and CE+CSDT (Section \ref{subsubsec:csdt}) losses presented in \cite{Cruz_2025}, using the Dice coefficient metric (\%). Again, higher values indicate better performance.}
    \label{tab:structured_loss_comparison_Dice}
    \scalebox{0.75}{
    \begin{tabular}{lcccc}
        \toprule
        Dataset 
        & CE 
        & CE + CSNP 
        & CE + CSDT
        & CE + CSSDF \\
        \midrule
        Breast Aesthetics & $93.8 \pm 0.5$ & \textcolor{blue}{$94.0 \pm 0.5$} & $94.0 \pm 0.4$ & $93.7 \pm 0.8$ \\
        Cervix-MobileODT  & $77.0 \pm 0.7$ & \textcolor{blue}{$77.4 \pm 0.2$} & $76.9 \pm 0.7$ & $76.8 \pm 0.3$ \\
        Mobbio   & $93.8 \pm 0.1$& $93.8 \pm 0.0$ & $93.7 \pm 0.2$& \textcolor{blue}{$93.8 \pm 0.1$}  \\
        Teeth-ISBI        & $74.0 \pm 1.6$ & $74.9 \pm 1.4$ & $75.3 \pm 1.3$ & \textcolor{blue}{$75.3 \pm 1.5$} \\
        Teeth-UCV         & $90.2 \pm 0.5$& \textcolor{blue}{$90.5 \pm 0.3 $}&$90.3 \pm 0.3$ &  $90.1 \pm 0.1$\\
        \bottomrule
    \end{tabular}}
\end{table}

\begin{table}[H]
    \centering
    \caption{Comparison of the structured loss functions investigated in this work with the CE+CSNP (Section \ref{subsubsec:contact_surface_metric}) and CE+CSDT (Section \ref{subsubsec:csdt}) losses presented in \cite{Cruz_2025}, using the unimodal pixels metric (\%). Again, higher values indicate better performance.}
    \label{tab:structured_loss_comparison_UP}
    \scalebox{0.75}{
    \begin{tabular}{lcccc}
        \toprule
        Dataset 
        & CE 
        & CE + CSNP 
        & CE + CSDT
        & CE + CSSDF \\
        \midrule
        Breast Aesthetics & $6.2 \pm 0.7$ & $5.9 \pm 2.1$ & $6.2 \pm 1.5$ & \textcolor{blue}{$7.0 \pm 1.2$} \\
        Cervix-MobileODT  & \textcolor{blue}{$1.0 \pm 0.2$} & $0.5 \pm 0.2$ & $0.9 \pm 0.3$ & $0.8 \pm 0.2$ \\
        Mobbio   & $0.8 \pm 0.1$ & \textcolor{blue}{$0.9 \pm 0.2$} & $0.7 \pm 0.2$ & $0.7 \pm 0.3$\\
        Teeth-ISBI        & $9.6 \pm 3.4$ & $11.2 \pm 4.6$ & $13.5 \pm 3.4$ & \textcolor{blue}{$13.7 \pm 3.5$}  \\
        Teeth-UCV         & $18.0 \pm 1.8$ & \textcolor{blue}{$25.7 \pm 6.8$} & $19.7 \pm 4.1$ & $20.1 \pm 5.0$ \\
        \bottomrule
    \end{tabular}}
\end{table}

Tables~\ref{tab:structured_loss_comparison_CS}–\ref{tab:structured_loss_comparison_UP} analyze the impact of structured ordinal loss functions. In Table~\ref{tab:structured_loss_comparison_CS}, CE+CSNP and CE+CSSDF achieve the lowest contact surface values across most datasets, demonstrating the effectiveness of spatial constraints in reducing boundary irregularities. The Dice results in Table~\ref{tab:structured_loss_comparison_Dice} indicate modest but consistent improvements over CE, with CE+CSNP achieving the best performance in most cases. Moreover, CE+CSSDF stands out compared to CE+CSDT, showing advantages in specific datasets. Finally, the unimodal pixels metric (Table~\ref{tab:structured_loss_comparison_UP}) shows that CE+CSSDF and CE+CSNP provide the highest pixel-level ordinal consistency, contributing to more coherent segmentations and smoothing irregularities along the boundaries of segmented regions.

Overall, these results indicate that unimodal and structured ordinal losses offer complementary benefits. While unimodal losses such as QUL and EXP\_MSE are more effective at improving structural consistency and Dice performance, structured losses enhance pixel-level ordinal coherence. This highlights the importance of jointly considering representation, structure, and segmentation accuracy when designing loss functions for ordinal semantic segmentation of medical images.

\section{Conclusion}
\label{sec:conclusion}
In this work, we investigated the impact of different unimodal and structured ordinal loss functions on the performance of a U-Net architecture applied to ordinal semantic segmentation of medical images. The experimental evaluation was conducted using 5-fold cross-validation across five datasets with distinct anatomical characteristics, enabling a comprehensive analysis of the influence of these loss functions from multiple perspectives, including structural consistency, pixel-level ordinal coherence, and segmentation accuracy.

The results demonstrate that incorporating unimodal losses, particularly QUL and EXP\_MSE, leads to consistent improvements in the contact surface metric, effectively reducing structural discontinuities between adjacent regions when compared to the conventional cross-entropy loss. Moreover, these losses maintain or improve the Dice coefficient across most datasets, indicating that enforcing ordinal constraints during training does not compromise segmentation quality and, in several cases, contributes to more accurate delineation of anatomical structures.

The analysis of the unimodal pixels metric shows that loss functions with strong pixel-level unimodality enforcement, such as CE+O2 and CE+QUL, achieve superior performance under this criterion. However, these improvements are not consistently reflected in global segmentation metrics, indicating that pixel-level ordinal coherence alone is insufficient to guarantee spatial consistency or overall segmentation performance.

Structured loss functions, in turn, prove effective in reducing boundary irregularities, particularly with respect to the contact surface metric, with CE+CSNP and CE+CSSDF achieving the lowest values in datasets with higher structural complexity. Although the improvements in the Dice coefficient are more modest, these losses contribute to enhanced ordinal coherence and improved spatial regularity in the resulting segmentations.

Overall, the findings indicate that unimodal and structured ordinal losses provide complementary benefits for ordinal semantic segmentation. While unimodal losses primarily favor the preservation of ordinal relationships and segmentation accuracy, structured losses reinforce spatial consistency and boundary regularity. These results highlight the importance of jointly considering representation, structure, and spatial correspondence when designing loss functions for ordinal segmentation tasks in medical imaging.

\bibliography{sn-bibliography}

\end{document}